\documentclass[sigconf,authorversion,nonacm]{acmart}

\copyrightyear{2021}
\acmYear{2021}
\setcopyright{acmlicensed}\acmConference[CIKM '21]{Proceedings of the 30th ACM International Conference on Information and Knowledge Management}{November 1--5, 2021}{Virtual Event, QLD, Australia}
\acmBooktitle{Proceedings of the 30th ACM International Conference on Information and Knowledge Management (CIKM '21), November 1--5, 2021, Virtual Event, QLD, Australia}
\acmPrice{15.00}
\acmDOI{10.1145/3459637.3482211}
\acmISBN{978-1-4503-8446-9/21/11}

\usepackage{amsthm}
\usepackage{mathtools}
\usepackage{graphicx}
\usepackage{wrapfig}
\usepackage{multirow}
\usepackage{subcaption}
\usepackage{algorithm}
\usepackage{algorithmic}
\usepackage{booktabs}

\begin{document}
\fancyhead{}



\title{Using Neighborhood Context to Improve Information Extraction from Visual Documents Captured on Mobile Phones}

\author{Kalpa Gunaratna, Vijay Srinivasan, Sandeep Nama, Hongxia Jin}
\affiliation{Samsung Research America, Mountain View CA \country{USA}}
\email{{k.gunaratna, v.srinivasan, s.nama, hongxia.jin}@samsung.com}


\begin{abstract}
Information Extraction from visual documents enables convenient and intelligent assistance to end users. We present a Neighborhood-based Information Extraction (NIE) approach that uses contextual language models and pays attention to the local neighborhood context in the visual documents to improve information extraction accuracy. We collect two different visual document datasets and show that our approach outperforms the state-of-the-art global context-based IE technique. In fact, NIE outperforms existing approaches in both small and large model sizes. Our on-device implementation of NIE on a mobile platform that generally requires small models showcases NIE's usefulness in practical real-world applications.
\end{abstract}




\maketitle

\section{Introduction}

\begin{figure}
    \centering
    \includegraphics[scale=0.42, trim=0cm 8.3cm 5.6cm 0cm, clip=true]{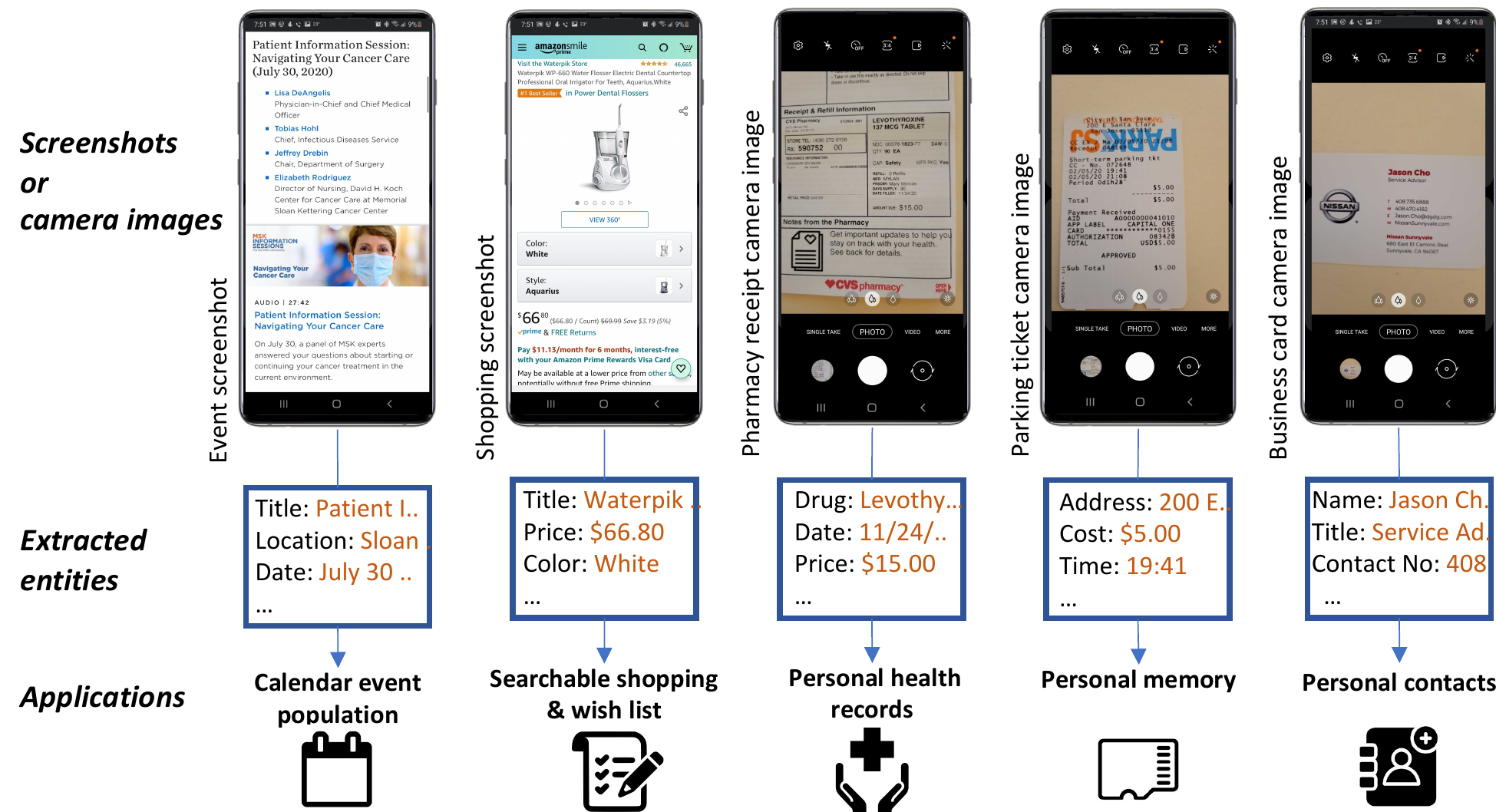}
    \caption{Applications of visual document IE on a mobile.}
    \label{fig_introduction}
\end{figure}

Information Extraction (IE)~\cite{cowie1996information} is the process of extracting structured information from unstructured or semi-structured documents. Though IE from text has been explored extensively in the literature, IE from visual documents such as event posters, receipts, product pages, tickets and medical bills as shown in Figure~\ref{fig_introduction} are comparatively less studied. Instead of the user manually entering such data, information extracted from such documents can be automatically entered to the user's personal calendar, shopping list, health record, or personal knowledge bases for convenient recall or processing by downstream applications.

Visual documents have distinct visual structures (text placement, font size, color) embedded with short text spans and are hence difficult to understand by only processing text, even for humans. While traditional text documents have grammatically correct long sentences, visual documents have short text phrases with visual features playing an important role. Thus, IE from visual documents requires different techniques compared to IE from text documents. Therefore, specific approaches (e.g., ~\cite{apostolova2014combining,palm2017cloudscan,liu2019graph,wei2020robust,qian2019graphie}) for this task have been introduced that consider both presentation structure and text. Rules and templates were popular techniques early on ~\cite{rusinol2013field, esser2013information, d2018field, sarkhel2019visual} but they are less flexible in modeling complex and challenging documents. Recent works proposed to capture the global document context~\cite{liu2019graph, wei2020robust} and document layout ~\cite{katti2018chargrid, yang2017learning}, where ~\cite{xu2020layoutlm} pre-trains on hundreds of thousands of document images to learn the layouts similar to the processing of contextual language models. Graph-based ones ~\cite{liu2019graph, wei2020robust} represent the visual segments of a document as a graph and capture the global context using graph context capturing techniques like graph convolutions~\cite{kipf2016semi}. However, since the edges between two visual segments are not well-defined, there is room for improvement over the graph-based global context capturing.

In this work, we present a Neighborhood-based IE (NIE) approach that uses contextual language models and pays attention to the local neighborhood of each text block (i.e., `blocks' identified by Optical Character Recognition - OCR) in the visual document to improve IE accuracy. Our intuition is that to extract information from a target text block, it is important to pay attention to neighboring text blocks which contain important hints about the entities in the target block. We collect two different visual document datasets to show that our approach outperforms the state-of-the-art global context-based IE technique for visual documents. Moreover, NIE outperforms existing best approaches for both small and large model sizes. We show the importance of achieving high accuracy in small models by considering a mobile phone use case where deployment of large models is not practical due to resource constraints. To the best of our knowledge, this is the first on-device IE solution for visual documents. Our contributions are as follows: (i) We propose a neighborhood context augmented contextual language model approach to improve visual document IE accuracy. (ii) We evaluate our approach on two domains of visual documents and show improvements over state-of-the-art visual document IE approaches across different model sizes. (iii) We implement our approach to perform IE from visual documents completely on the mobile phone itself to demonstrate the importance of improving accuracy on smaller model sizes that can also preserve user privacy.

\section{Approach}

\paragraph{\textbf{Problem:}} Given a visual document $D$ with less syntactically structured text and rich visual features as input, we identify the set of entity spans $\{E\}$ in $D$, each belonging to one of the predefined classes in set $\{K\}$. Each $e \in \{E\}$ is a sub sequence in $D_{text}$ and $Type(e) \in {K}$, where $D_{text}$ is obtained by applying OCR on $D$. 

\begin{figure}
    \centering
    \includegraphics[scale=0.6, trim=0.1cm 0.75cm 14cm 3.8cm, clip=true]{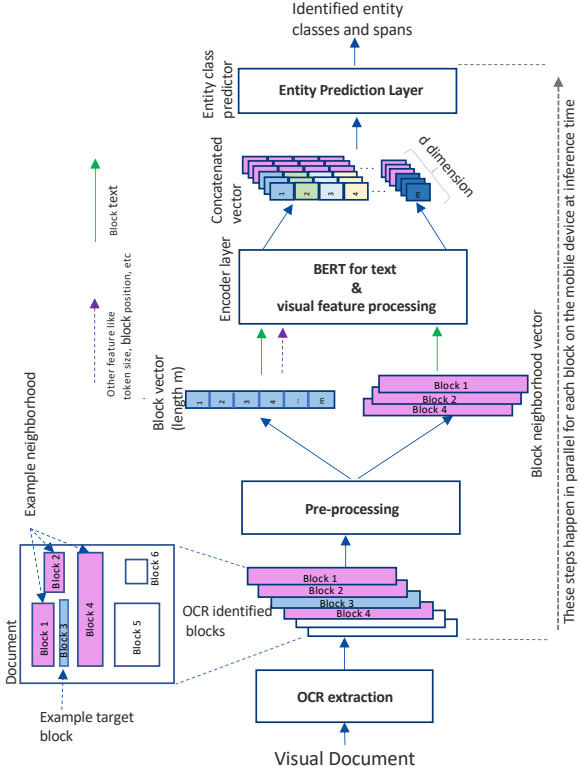}
    \caption{The processing steps overview using BERT with respect to one block (blue) selected from the OCR output.}
    \label{fig_approach}
\end{figure}

\paragraph{\textbf{Overview:}}The overview of our approach is illustrated in Figure~\ref{fig_approach}. We compute the local neighborhood context for each text block (identified by the OCR) and use this local neighborhood context in contextual language models like BERT~\cite{devlin2019bert} to predict the entity class labels (in IOB notation) for the input tokens in each block. Since we use an off the shelf OCR algorithm, one large chunk of content in a document that should be considered as one piece can be accidentally split into multiple blocks. To reduce this effect, we merge two adjacent blocks if they are closer than a predefined threshold. In summary: (i) we use the BERT contextual language model to compute embeddings for each token (word) of the text in the target processing block (i.e., the block that we want to predict entity classes), (ii) we use the same BERT model to compute a neighborhood context embedding by processing nearby blocks, and (iii) we compute visual features for each token in the target text block based on OCR output. The above three signals are combined to predict the entity class for each token in the text of the target processing block.

\paragraph{\textbf{Local Neighborhood Context:}}
Our approach uses the local neighborhood of each block in contrast to capturing the whole document (i.e., global) context as in the state-of-the-art~\cite{wei2020robust}. The local context for the $i$th block $b^{i}$ of a block set $\{b^{1}, b^{2}, .., b^{N}\}$ belonging to a document $D$ can be captured in three different ways using a sliding window approach: (i) Top neighborhood context: captures content of $n$ blocks appearing before the $i$th block ($\{b^{i-n}, b^{i-n+1}, .., b^{i-1}\}$), (ii) Bottom neighborhood context: captures content of $n$ blocks appearing after the $i$th block ($\{b^{i+1}, b^{i+2}, b^{i+n}\}$), and (iii) Overlapping neighborhood context: total of $n$ blocks from both before and after the $i$th block are considered for capturing the context. If $n$ is even, we take $n/2$ blocks each from above and below whereas, when $n$ is odd, we take $(n+1)/2$ from the above and $n - (n+1)/2$ blocks following the $i$th block.

\paragraph{\textbf{Entity Prediction Model:}}
Given a block $b^{i}$ with a list of tokenized words $[tok^{i}_{1}, tok^{i}_{2}, tok^{i}_{3}, .. , tok^{i}_{k}]$, we use the list of tokenized words as the input to BERT model to get the block embeddings $T^{i}_{0:k}$ where, $tok^{i}_{0}$ is the special token $[CLS]$ that is used to get the aggregate context representation ($T^{i}_{0}$) of the input text block $b^{i}$. The context representation $C^{i}$ of block $b^{i}$ is $T^{i}_{0}$, the embedding for the CLS token. Note that any language model can be used as the encoder, even though we use BERT in our implementation.

\begin{equation}
    T^{i}_{0:k} = BERT(tok^{i}_{0:k};\theta)
\label{eq:bert}
\end{equation}

The neighborhood context for a block may be computed in one of three context capturing ways mentioned above. We concatenate the neighborhood block content and use the BERT model to get the neighborhood context embedding. This neighborhood context embedding is then concatenated into each token BERT embedding of the current processing block. If there are $n$ neighborhood blocks $b^{1}, b^{2}, .., b^{n}$ identified for block $b^{j}$, which is the target processing block, then the neighborhood text $N^{j}$ of block $b^{j}$ is computed by appending the tokens in the neighboring blocks in the order they appear in the document as [$b^{1} || b^{2} || .. || b^{n}$], where $||$ represents appending two lists. Then applying $N^{j}$ in Equation~\ref{eq:bert}, we get the neighborhood context vector $C^{j}$ as the embedding for the CLS token. Our neighborhood capturing uses a simple appending mechanism and hence can be efficiently computed, even in resource constrained mobile phone platforms. We show in our empirical evaluation that our local context capturing works well to improve IE accuracy across small and large model sizes.

The neighborhood context $C^{j}$ is then concatenated ($\oplus$) to each token embedding $T^{j}_{x}$ of token $tok^{j}_{x}$ in the target processing block $b^{j}$. This neighborhood content is expected to provide nearby block context for better prediction of each token's entity class. Additionally, visual features ($f$) such as token size and block location are concatenated to the token embeddings. Visual features provide additional spatial and style signals to the model; for example, title of an event poster may appear in top part of the document possibly with larger font size. These features are projected into the embedding space and concatenated to the target processing block's token embeddings to get the final token embedding $V^{j}_{x}$ as follows. 

\begin{equation}
    V^{j}_{x} = T^{j}_{x} \oplus C^{j} \oplus f^{j}_{x}
\label{eq:block_concat}
\end{equation}

where, $T^{j}_{x}$ $\in$ $T^{j}_{1:k}$, $T^{j}_{x} \in \mathbb{R}^{d1}$, $C^{j} \in \mathbb{R}^{d2}$, $f^{j}_{x} \in \mathbb{R}^{d3}$, and $V^{j}_{x} \in \mathbb{R}^{d1+d2+d3}$. 

We predict the entity classes using a fully connected linear layer.

\begin{equation}
    predicted\_class = arg max(V^{j}_{x}W + b)
\label{eq:classify}
\end{equation}

where, $W \in \mathbb{R}^{(d1+d2+d3) \times l}$, $l$ is number of distinct entity classes to predict in the IOB tagging scheme~\footnote{If there are $\eta$ main classes where $\eta$=$|K|$, there will be $\eta \times 2 + 1$ classes in IOB.}.

\section{Evaluation}
Due to the unavailability of public visual document datasets and especially ones that operate on mobile screen sized images/documents to support our use case implementation, we created two datasets: (i) event poster/webpage screenshots and (ii) product webpage screenshots with product details. These two datasets, event posters and product details facilitate two probable real-world use cases, calendar event population and product (i.e., shopping) wish list population and search, respectively, in a mobile personal assistant environment (see Figure~\ref{fig_introduction}). The dataset statistics are shown in Table~\ref{tab:datasets}. The datasets were annotated using a crowd-sourcing platform with multiple annotations per entity span. We kept documents with entity annotations that received more than 50\% agreement (having at least 7 judgements per annotation) and removed others to maintain high quality entity annotations. We tag `title' and `price' entity spans for both datasets and additionally tag `location' and `time' spans for the event dataset. We evaluated our approach against two state-of-the-art baselines: (i) vanilla BERT that uses no context and (ii) graph convolutions-based (GCN) global context capturing approach~\cite{wei2020robust}. In the literature~\cite{wei2020robust}, the GCN baseline was shown to be the state of the art approach and contextual language models such as BERT outperformed sequence models (e.g., LSTMs) for this task . For NIE, we used bottom neighborhood context capturing method with $n$=4 to report results as it showed the best accuracy. We use token font size and y coordinates of the blocks as the visual features for all the models. We report micro-averaged precision ($Prec$), recall ($Rec$), and F1 score, where the named entity is considered correct when both boundary and type are predicted correctly. 

\begin{table}[]
\begin{tabular}{llllll}
\hline
Dataset                         & Total     & Removed      &Train      & Dev       & Test\\ \hline
Event                          & 3061      & 854           & 1544      & 331       & 332  \\ 
Product                         & 1646      & 122           & 1066      & 228       & 230  \\ \hline
\end{tabular}
\caption{Dataset statistics. Total is the total number of images collected and Removed ones have less than 50\% user tagging agreement (from at least 7 user annotations).}
\label{tab:datasets}
\end{table}

\paragraph{\textbf{Accuracy Across Model Sizes:}} First, we report an analysis of accuracy (F1) over different model sizes for the two datasets. The results are shown in Tables ~\ref{tab:event} and ~\ref{tab:store}. In both the datasets, our NIE approach outperforms the vanilla BERT (that uses no local or global context) and global context-based GCN approach. Interestingly, the GCN approach only shows improvements in the product dataset over the BERT baseline. This may be due to the fact that visual documents are hard to represent as a graph using block level distance or similar measures~\cite{wei2020robust}. Further, using this graph modeling to capture global context representation for the document may be too coarse grained and the model may not get much benefit from the global context. In contrast, our model can pay attention to the local neighborhood context to decide the entity spans in the target text block. For example, by doing so, the model can learn that `title' normally appears before the `location' block and also they are typically not in the neighborhood of the `price' information in event posters. From the two datasets, our model seems to have higher relative improvement in accuracy compared to the baselines in the product dataset. This may be due to the fact that, product descriptions have similar wording throughout the page. For example, `title' of a product description page has similar wording to its description. Hence, ability to closely monitor the local neighborhood, instead of no context (like in vanilla BERT) or global context (like in GCN) seems to be highly beneficial in identifying the product title correctly.

\begin{table}[]
\begin{tabular}{lllll}
\hline
Model Size                              & Method & F1              & Prec            & Rec        \\ \hline
\multirow{3}{*}{BERT Tiny (17MB)}       & BERT   & 0.7325          & 0.7017          & 0.7661          \\
                                        & GCN    & 0.7429          & 0.7175          & 0.7703          \\
                                        & NIE   & \textbf{0.7708} & \textbf{0.7459} & \textbf{0.7976} \\ \hline
\multirow{3}{*}{BERT Mini (45 MB)}      & BERT   & 0.7823          & 0.7709    & 0.7941          \\
                                        & GCN    & 0.7691          & 0.7506    & 0.7885          \\
                                        & NIE   & \textbf{0.7928} & \textbf{0.7756} & \textbf{0.8109} \\ \hline
\multirow{3}{*}{BERT Small (116 MB)}    & BERT   & 0.7834          & 0.7777    & 0.7892          \\
                                        & GCN    & 0.7719          & 0.7467    & 0.7990          \\
                                        & NIE   & \textbf{0.8149} & \textbf{0.8079} & \textbf{0.8221} \\ \hline
\multirow{3}{*}{BERT Medium (167 MB)}   & BERT   & 0.7905          & 0.7919    & 0.7892          \\
                                        & GCN    & 0.7862          & 0.7592    & 0.8151          \\
                                        & NIE   & \textbf{0.8148} & \textbf{0.7983} & \textbf{0.8319} \\ \hline
\multirow{3}{*}{BERT Base (440 MB)}     & BERT   & 0.8103          & 0.8069    & 0.8137          \\
                                        & GCN    & 0.8035          & 0.7923    & 0.8151          \\
                                        & NIE   & \textbf{0.8247} & \textbf{0.8111} & \textbf{0.8389} \\ \hline
\end{tabular}
\caption{Results for Event dataset. BERT used no context; GCN and our NIE approaches used GCN-based global context + BERT and neighborhood context + BERT, respectively. All methods used size and y coordinate features.}
\label{tab:event}
\end{table}

\begin{table}[]
\begin{tabular}{lllll}
\hline
Dataset                                 & Method & F1              & Prec      & Rec        \\ \hline
\multirow{3}{*}{BERT Tiny (17 MB)}      & BERT   & 0.6587          & 0.6387    & 0.6799          \\
                                        & GCN    & 0.5900          & 0.5080    & 0.7036          \\
                                        & NIE   & \textbf{0.7284} & \textbf{0.6979} & \textbf{0.7617} \\ \hline
\multirow{3}{*}{BERT Mini (45 MB)}      & BERT   & 0.6922          & 0.6684          & 0.7178          \\
                                        & GCN    & 0.8169          & 0.7837          & 0.8531          \\
                                        & NIE   & \textbf{0.8792} & \textbf{0.8617} & \textbf{0.8975} \\ \hline
\multirow{3}{*}{BERT Small (116 MB)}    & BERT   & 0.7052          & 0.6948          & 0.7159          \\
                                        & GCN    & 0.8470          & 0.8280          & 0.8670          \\
                                        & NIE   & \textbf{0.8858} & \textbf{0.8797} & \textbf{0.8919} \\ \hline
\multirow{3}{*}{BERT Medium (167 MB)}   & BERT   & 0.7158          & 0.7335          & 0.6988          \\
                                        & GCN    & 0.8389          & 0.8201          & 0.8587          \\
                                        & NIE   & \textbf{0.9038} & \textbf{0.8964} & \textbf{0.9113} \\ \hline
\multirow{3}{*}{BERT Base (440 MB)}  & BERT   & 0.7209          & 0.7380          & 0.7045          \\
                                        & GCN    & 0.7168          & 0.7203          & 0.7121          \\
                                        & NIE   & \textbf{0.7561} & \textbf{0.7605} & \textbf{0.7518} \\ \hline
\end{tabular}
\caption{Results for Product dataset. BERT used no context; GCN and our NIE approaches used GCN-based global context + BERT and neighborhood context + BERT, respectively. All methods used size and y coordinate features.}
\label{tab:store}
\end{table}

\begin{figure}
    \centering
    \includegraphics[scale=0.31, trim=0.2cm 12cm 0.2cm 0cm, clip=true]{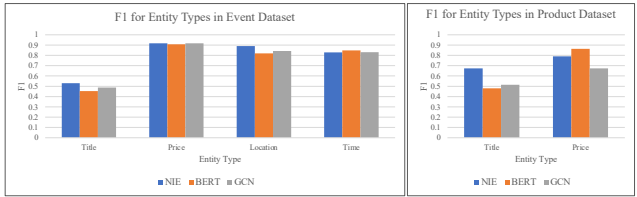}
    \caption{Class level F1 for the two datasets using the 17 MB model with size and y coordinate visual features.}
    \label{fig_type_f1}
\end{figure}

\paragraph{\textbf{Entity Class Level Accuracy:}} Due to space limitations, we illustrate the class level F1 breakdown for the two datasets for all the approaches using the smallest model size (17 MB) in Figure~\ref{fig_type_f1}. The breakdown is similar for all the other model sizes. In the breakdown, it is easy to see that all the models perform well for `price', `location', and `time' predictions. This is mainly because they are comparatively less complex to detect and are less ambiguous. In contrast, all the models struggle to achieve high F1 for `title' predictions. We see that our NIE approach outperforms other approaches by a considerable margin for the complex `title' class, as explained above. We also computed macro-averaged F1 (average class-level F1) to show that our accuracy is not unfairly affected by the improvement in one class such as `title'. For the event dataset, macro F1 scores are 0.7567, 7695, and 7920 for vanilla BERT, GCN, and NIE, respectively. For the product dataset, they are 0.6720, 0.5937, and 0.7316 for vanilla BERT, GCN, and NIE, respectively. This clearly reflects that NIE performs well across classes for both the datasets.

\paragraph{\textbf{Improvements from Visual Features:}} Incorporating visual features can improve base model accuracy. This is because the model gets additional hints from the visual cues that can help determine certain entities. For example, `title' in a poster is generally presented in a larger font and typically towards the top of the document (small y-coordinate). Table~\ref{tab:features} shows results of NIE with and without visual features, using only the 17MB model as a reference due to space limitations; the ablation results for other model sizes are similar.

\begin{table}[]
\begin{tabular}{lllll}
\hline
Dataset                                 & Use of Visual Features            & F1                & Prec              & Rec        \\ \hline
\multirow{3}{*}{Event}                  & NIE w/o features                 & 0.7616            & 0.7258            & \textbf{0.8011}          \\
                                        & NIE w features                   & \textbf{0.7708}   & \textbf{0.7459}   & 0.7976          \\ \hline
\multirow{3}{*}{Product}                & NIE w/o features                 & 0.6684            & 0.6438            & 0.6950          \\
                                        & NIE w features                   & \textbf{0.7284}   & \textbf{0.6979}   & \textbf{0.7617}          \\ \hline
\end{tabular}
\caption{Effect of visual features (size and y coordinates) on NIE using 17MB model. w/o - without, w - with.}
\label{tab:features}
\end{table}

\paragraph{\textbf{Mobile Use Case and Implementation:}} We have shown that our neighborhood-based approach outperforms the baselines across all the model sizes. In this use case, we show that we can implement and deploy a visual document IE system to completely process and infer the entity classes in the mobile device without connecting to an external server. This is extremely useful in preserving user privacy in mobile data processing (e.g., ~\cite{betzing2020impact}). For mobile apps, we need to use a model with small memory consumption and hence we use our smallest 17 MB model (NIE is the best performing model) in this use case experiment. In the Android app, document pre-processing is done using Java and the deep learning model inference was run using PyTorch Mobile after converting source code from PyTorch to TorchScript. The model inference time on the mobile device was on average about 60 milliseconds (measured on a Samsung Galaxy S20 by averaging 5 random documents). The importance of making small models achieve high accuracy is evident by looking at the 167 MB model, which took 1267 milliseconds for inference on the same device; this latency would result in a noticeable delay for the end user. The 440 MB model crashed while loading on the mobile, showing that large models are not suitable for mobile platforms.

For mobile deployments and integrating into a commercial mobile platform, it is important to keep models sizes in single digit mega bytes range, especially with regards to Figure~\ref{fig_introduction} where we may need to store specialized models for each sub domain to have high accuracy. We performed experiments to further reduce model sizes by applying training-based quantization techniques. The F1 results after applying 8-bit quantization (reduced from 32-bits) are presented in Table~\ref{tab:quantized_results}. The drop is noticeable only in the product dataset where some information may be lost in the 8-bit representation of the model, but NIE still outperformed all baselines. We were able to reduce 17 MB and 45 MB models to approximately 4.25 MB and 11.25 MB, respectively.


\begin{table}[]
\begin{tabular}{llrrrr}
\hline
                         &      & \multicolumn{2}{c}{BERT Tiny (17 MB)} & \multicolumn{2}{c}{BERT Mini (45 MB)} \\ \cline{3-6} 
                         &      & Original      & Quantized     & Original      & Quantized     \\ \hline
\multirow{3}{*}{Event}   & BERT & 0.7325        & 0.7258        & 0.7823        & 0.7777        \\
                         & GCN  & 0.7429        & 0.7447        & 0.7691        & 0.7586        \\
                         & NIE & 0.7708        & 0.7606        & 0.7928        & 0.8008        \\ \hline
\multirow{3}{*}{Product} & BERT & 0.6587        & 0.6222        & 0.6922        & 0.6948        \\
                         & GCN  & 0.5900        & 0.6172        & 0.8169        & 0.6841        \\
                         & NIE & 0.7284        & 0.6511        & 0.8792        & 0.7429        \\ \hline
\end{tabular}
\caption{Original Vs. 8-bit quantized model F1 scores for the two smaller models.}
\label{tab:quantized_results}
\end{table}

\section{Conclusion}
We proposed and evaluated NIE, a local neighborhood context-based contextual language model for visual document IE. NIE outperformed state-of-the-art methods in accuracy across model sizes. Additionally, our complete on-device mobile implementation showcased the potential of NIE to enable a privacy preserving intelligent and \textit{personal} user assistance. We plan to investigate more on neighborhood processing techniques to handle more complex visual documents with multiple events or products, further improve accuracy while maintaining low model complexity, and explore more on knowledge distillation~\cite{gou2021knowledge} and quantization~\cite{guo2018survey} techniques.

\bibliographystyle{ACM-Reference-Format}
\bibliography{bibliography}

\end{document}